\documentclass[runningheads]{llncs}

\usepackage{graphicx}

\usepackage{tikz}
\usepackage{comment}
\usepackage{amsmath,amssymb} 
\usepackage{color}
\usepackage{multirow}
\usepackage{booktabs}
\usepackage{subcaption}
\usepackage{bm}
\usepackage{array}
\usepackage{xcolor}
\usepackage{comment}
\usepackage{amsopn}
\usepackage{xspace}
\usepackage{hyperref}

\makeatletter
\DeclareRobustCommand\onedot{\futurelet\@let@token\@onedot}
\def\@onedot{\ifx\@let@token.\else.\null\fi\xspace}

\def\eg{\emph{e.g}\onedot} 
\def\ie{\emph{i.e}\onedot}

\def\etal{\emph{et al}\onedot}
\makeatother

\begin{document}
\pagestyle{headings}
\mainmatter
\def\ECCVSubNumber{7027}  

\title{Supervised Attribute Information Removal and Reconstruction for Image Manipulation } 

\titlerunning{AIRR}
%
\author{Nannan Li \and
Bryan A. Plummer}
\authorrunning{N. Li, B. A. Plummer}
%
\institute{Boston University\\
\email{\{nnli,bplum\}@bu.edu}}
\maketitle

\begin{abstract}
The goal of attribute manipulation is to control specified attribute(s) in given images. Prior work approaches this problem by learning disentangled representations for each attribute that enables it to manipulate the encoded source attributes to the target attributes. However, encoded attributes are often correlated with relevant image content.  Thus, the source attribute information can often be hidden in the disentangled features, leading to unwanted image editing effects. In this paper, we propose an Attribute Information Removal and Reconstruction (AIRR) network that prevents such information hiding by learning how to remove the attribute information entirely, creating attribute excluded features, and then learns to directly inject the desired attributes in a reconstructed image. We evaluate our approach on four diverse datasets with a variety of attributes including DeepFashion Synthesis, DeepFashion Fine-grained Attribute, CelebA and CelebA-HQ, where our model improves attribute manipulation accuracy and top-k retrieval rate by 10\% on average over prior work. A user study also reports that AIRR manipulated images are preferred over prior work in up to 76\% of cases\footnote{Code and models are available at \url{https://github.com/NannanLi999/AIRR}}.
\end{abstract}

\section{Introduction}
Attribute manipulation translates images based on desired attributes, which has applications to face editing \cite{shen2020interpreting,Wu_2021_CVPR,wang2021attribute}, image retrieval \cite{shin2019semi,hou2021learning,yang2020generative}, and image synthesis \cite{ak2019attribute,kwon2022tailor}, among others. In these tasks, the goal is to be able to control a specified attribute without affecting other information in the source image. While Generative Adversarial Networks (GANs) have achieved impressive performance on attribute manipulation, a major challenge is that the generator tends to take a shortcut by utilizing the preserved source attribute information instead of the target attribute for manipulation \cite{hu2018disentangling,szabo2018understanding,lezama2018overcoming}, thus causing improper image editing effects in manipulated images. Prior work has tried to address this by adding random noise during the reconstruction \cite{bashkirova2019adversarial,usman2021disentangled} or learning disentangled attribute representations which are used to manipulate the images \cite{shen2020interpreting,yang2021l2m}. However, low-magnitude random noise is not targeted to the source attribute, which could be intentionally ignored by the model or inadvertently suppress key source features. On the other hand, even with a disentangled image representation, the correlation between attributes and relevant image content could cause source attribute information to be hidden in the rest of the image features. For example, attribute \emph{formal} in a dress is often correlated with the dress's \emph{long length}.

\begin{figure*}[!t]
    \centering
    \begin{subfigure}[c]{\textwidth}
  \centering
    \includegraphics[width=0.9\textwidth]{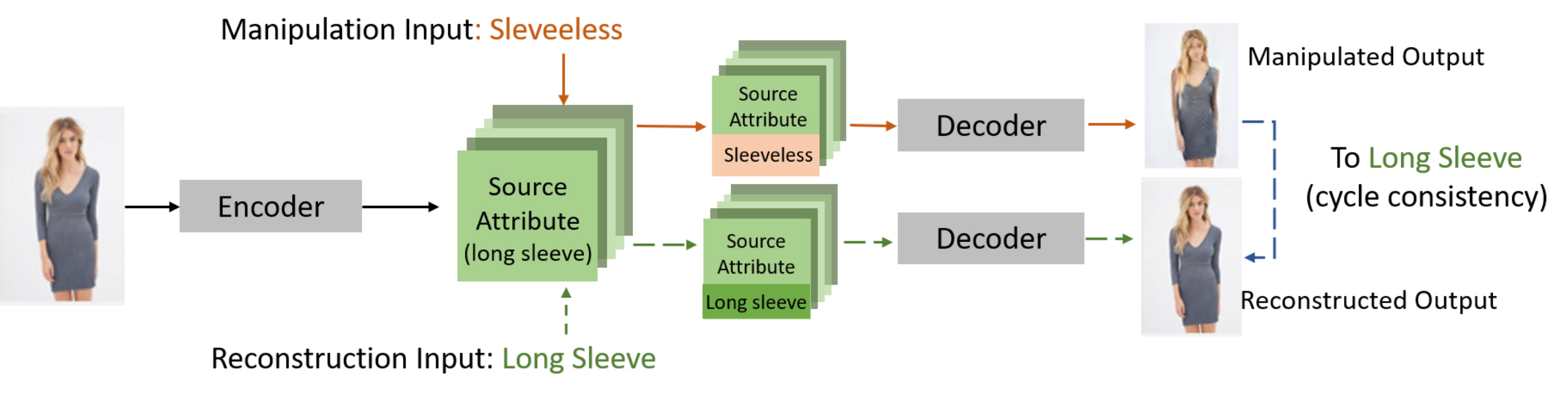}
    \caption{Framework of previous methods \cite{ak2019attribute,hou2021learning,yang2020generative,yao2021latent,he2019attgan}. Dashed arrows mean two different ways of obtaining the reconstructed image}
 \end{subfigure}
  \begin{subfigure}[c]{\textwidth}
  \centering
    \includegraphics[width=\textwidth]{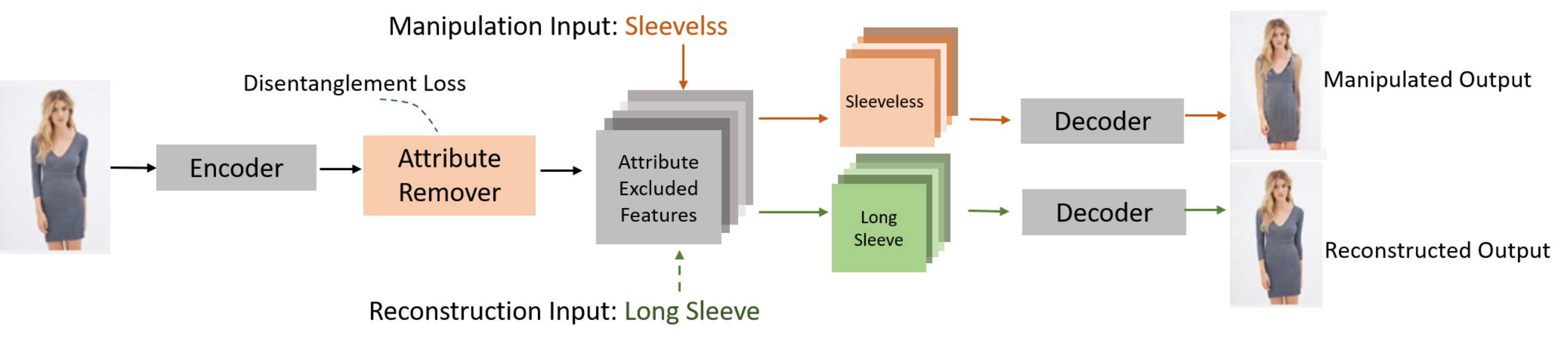}
    \caption{Pipeline of the proposed method}
   
 \end{subfigure}
    \caption{In (a), the generator like those used by \cite{he2019attgan,ak2019attribute,hou2021learning,yang2020generative,yao2021latent} incorrectly utilizes the hidden source attribute information instead of the target attribute \emph{sleeveless} for image manipulation. As a result, the manipulated image still contains the source attribute \emph{long sleeve}, causing improper image editing effects. To avoid this issue, the proposed method in (b) erases the source attribute information in the encoded features through an attribute remover with a disentanglement loss, conditioning the manipulated output only on the input target attribute \emph{sleeveless} and the attribute excluded features}
 \label{fig:motiv}
\end{figure*}

To address these issues, we propose a supervised Attribute Information Removal and Reconstruction (AIRR) model that learns an attribute excluded representation and reconstructs the image with desired attributes. The key challenge is in identifying the preserved source attribute information and decorrelating it from the image representation \cite{kwak2020cafe,wang2021attribute,shen2020interpreting,yang2021l2m}. Prior research on feature disentanglement either doesn't consider the decorrelation \cite{kwak2020cafe,wang2021attribute,yao2021latent}, or has limitations on the number of attributes it can decorrelate in a forward pass \cite{shen2020interpreting,yang2021l2m}, unlike our approach which can disentangle any number of attributes. 
In addition, as illustrated in Figure \ref{fig:motiv}a, these methods often rely on the full image information for both manipulation and reconstruction. As mentioned earlier, this can lead to information hiding in the manipulated image. In contrast, as shown in Figure \ref{fig:motiv}b, we use our remover to erase attribute information to obtain attribute excluded features, which are then used to directly generate both the reconstructed and manipulated images. Since this should eliminate the information that could potentially be hidden in the disentangled representation, we avoid the information hiding issues in prior work.

One challenge in our approach is our reliance on being able to identify and remove attribute information in real images.  For example, although the color \emph{white} appears in the background of the input images in Figure \ref{fig:motiv}, a good attribute classifier would predict that the clothing item is \emph{gray} and not \emph{white}. This means that the background information could mislead the attribute recognition. To address this issue, we segment the object of interest (\eg, using \cite{li2020self,yu2018bisenet}) to split the image encoder into two branches for the object of interest and the background, respectively. This helps AIRR to concentrate the manipulation on the object of interest without influencing the background information.

Our main contributions are:
\begin{itemize}
    \item We propose the Attribute Information Removal and Reconstruction method (AIRR), a controllable disentangled attribute manipulation framework that produces high quality images. The key insight in AIRR is the attribute information removal and reconstruction module that produces an attribute excluded representation, eliminating sources of information hiding that degrades performance in prior work.
    \item Extensive experiments across DeepFashion Synthesis~\cite{liuLQWTcvpr16DeepFashion}, DeepFashion Fine-grained Attribute~\cite{liuLQWTcvpr16DeepFashion}, CelebA~\cite{liu2015faceattributes} and CelebA-HQ~\cite{CelebAMask-HQ} report that AIRR improves the attribute manipulation accuracy and top-k retrieval rate by 10\% on average over the state-of-the-art. Moreover, we show that AIRR can effectively control attribute strength as well as efficiently manipulating multiple attributes in a single forward pass.
    \item A user study further validates the effectiveness of our approach, where our methods are shown to produce high quality images that more accurately achieve the target attribute manipulation by up to 76\% over prior work.
\end{itemize}

\section{Related Work}

Early research in attribute manipulation \cite{ChoiCKH0C18,he2019attgan} combined the target attribute label directly with the image or image features, and decoded them into manipulated output. However, the decoder could incorrectly use the preserved source attribute information for image manipulation. Thus, more recent work (including this paper), has focused on learning disentangled attribute representations, which we will discuss in more detailed below.

\noindent
\textbf{Unsupervised disentanglement.} Several studies explored disentanglement in the latent space of GANs in an unsupervised manner \cite{ramesh2018spectral,locatello2019challenging,harkonen2020ganspace,shoshan2021gan,Wu_2021_CVPR}. These methods aim to manipulate the attributes on synthetic data, where the image content is randomly generated. In \cite{harkonen2020ganspace}, the authors found that the principle components of features on pretrained GANs represent high-level semantic concepts. In \cite{Wu_2021_CVPR}, the authors introduced channel-wise disentanglement of StyleGAN \cite{karras2019style}. Shoshan~\etal \cite{shoshan2021gan} utilized contrastive learning to disentangle the latent space, achieving explicit control over synthetic facial images. However, without manual examinations on the feature space, it's difficult to locate the exact attribute representation that we want to manipulate, especially for attributes with high-level semantics.  Thus, in our work we focus on cases where attributes we wish to manipulate are known, enabling us to directly target our feature learning.
\smallskip

\noindent
\textbf{Supervised disentanglement.} Supervised disentanglement methods edit real images based on attribute annotations. Prior work on this task can be categorized in two types: spatial disentanglement and feature disentanglement. In methods that focus on spatial disentanglement \cite{kwak2020cafe,wang2021attribute}, attributes are located spatially and thus disentangled in the feature map. These methods can find attribute-specific features by an attention map, whereas  attribute-relevant information is implicitly kept and thus influences the image manipulation. On the other hand, feature disentanglement identifies certain features corresponding to the manipulated attribute. \cite{yao2021latent} presents a method that learns a linear transformation function that maps StyleGAN's latent code. Although StyleGAN's latent space is disentangled \cite{abdal2020image2stylegan}, without orthogonal constraints, such linear combination could result in correlation between different image attributes and content. To address this, Yang~\etal \cite{yang2021l2m} learned attribute relevant and irrelevant features, but each manipulated attribute requires training its own model, which is computationally costly. Instead, Shen~\etal~\cite{shen2020interpreting} manipulated attributes with a conditional subspace projection via Support Vector Machines (SVM), whereas the manipulation accuracy depends on the capability of SVM and each forward pass can control only a single attribute. In contrast, our proposed approach can manipulate multiple source attributes in a single forward pass by utilizing the injected attribute embedding. 
\smallskip

\noindent
\textbf{Attribute manipulation in fashion.} Apart from the above mentioned methods that are mainly applied to facial attribute editing, attribute manipulation in fashion images has also gained a lot of attention. Recent work on this topic mainly aim to improve the image retrieval accuracy for item recommendation. For example, researches have leveraged spatial information when manipulating attributes~\cite{ak2019attribute,ak2021fashionsearchnet}, learned a dictionary of attribute transformations \cite{shin2019semi}, or used the attribute probability distribution as an disentangled representation for image retrieval\cite{hou2021learning}. Kwon~\etal~\cite{kwon2022tailor} predicted changes to an item's shape as a result of changing an attribute, enabling them to make more significant alterations to the clothing in images. However, many of these methods also suffered from issues with disentangling attributes, often due to misinformation hiding, which our work minimizes.

\begin{figure*}[!t]
\centering
    \includegraphics[width=\textwidth]{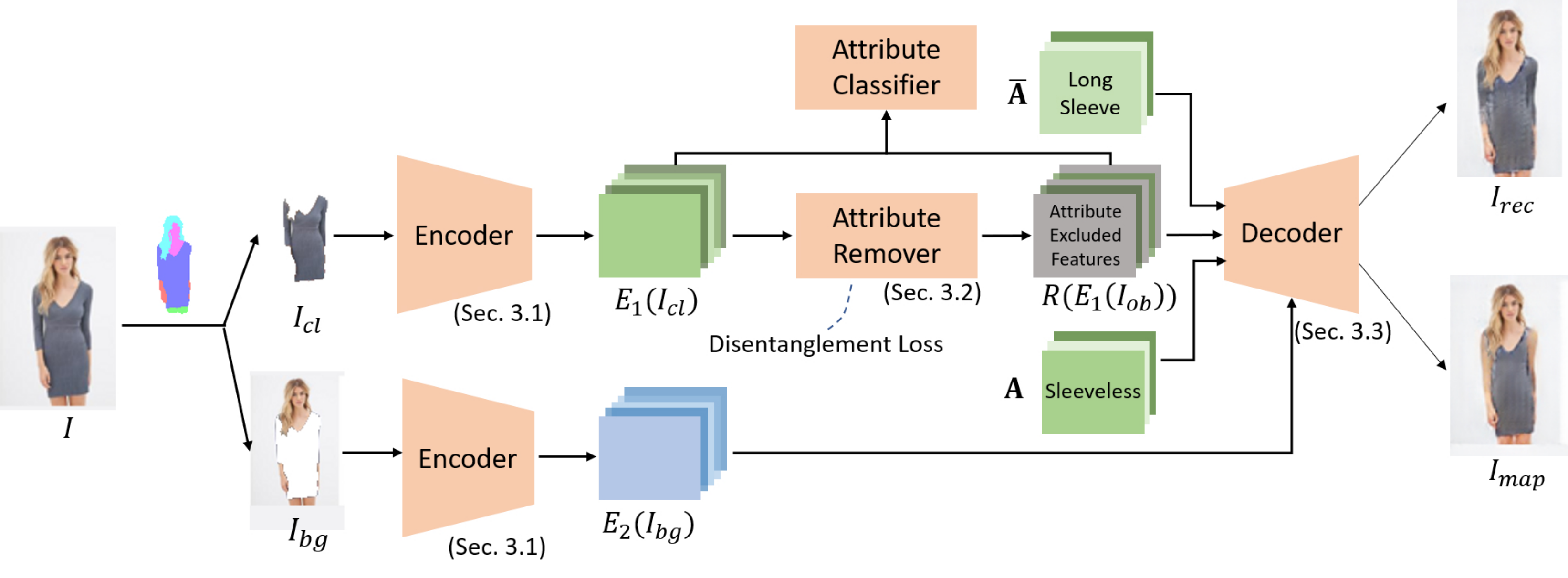}
  \caption{\textbf{AIRR framework}. In the AIRR generator, a given image is parsed into an object of interest $I_{cl}$ and background $I_{bg}$ through an offline parser \cite{li2020self,yu2018bisenet}. $I_{cl}$ and $I_{bg}$ are encoded in separate branches (Sec.\ \ref{sec:encoder}). In the $I_{cl}$ branch, the source attribute information in the encoded features are erased by an attribute remover using a disentanglement loss (Sec.\ \ref{sec:remover}). Subsequently, the source attributes in $\mathbf{A}$ and the target attributes in $\mathbf{\bar{A}}$ are embedded into the attribute excluded features for image reconstruction and image manipulation, respectively (Sec.\ \ref{sec:decoder})} 
  \label{fig:method}
\end{figure*}

\section{Attribute Information Removal and Reconstruction}
Given image $I$ and its attributes $\mathbf{A} = \{ \mathbf{a_1},\mathbf{a_2},...,\mathbf{a_n}\}$, where $a_i$ denotes the $i$th attribute, we aim to manipulate any number of attributes in $\mathbf{A}$. To achieve this goal, the generator first takes a real image as input, and uses our attribute remover to decorrelate the image attributes from the image features. The resulting attribute excluded representation is then combined with the target attribute embeddings to produce the manipulated output $I_{map}$. In the following, we introduce the four components of our model: image encoder (Section \ref{sec:encoder}), attribute remover (Section \ref{sec:remover}), decoder (Section \ref{sec:decoder}), and learning objectives (Section \ref{sec:other_loss}). Figure \ref{fig:method} shows an overview of our approach.  

\subsection{Image Encoder}
\label{sec:encoder}

To concentrate the manipulation on the object that we want to manipulate, the image encoder in AIRR is split into two branches for the object of interest $I_{cl}$ and the background $I_{bg}$, respectively. Prior work often achieves the segmentation of $I_{cl}$ by learning an attention map \cite{kwak2020cafe,ak2019attribute,ak2021fashionsearchnet}, while we found empirically that using an offline parser \cite{li2020self,yu2018bisenet} is more accurate in segmenting instances. This segmentation is especially helpful for images with multiple objects, \eg, an image of a fashion model wearing top, leggings and boots. As shown in Figure \ref{fig:method}, after obtaining $I_{cl}$ and $I_{bg}$, the image encoder encodes $I_{cl}$ into $E_1(I_{cl})$ in the first branch, and $I_{bg}$ into $E_2(I_{bg})$ in the second branch. Later on, AIRR only manipulates $E_1(I_{cl})$ without influencing the background information $E_2(I_{bg})$.

\subsection{Attribute Remover} 
\label{sec:remover}
While prior work directly used the image features $E_1(I_{cl})$ to generate the image \cite{he2019attgan,ak2019attribute,hou2021learning,yang2020generative,yao2021latent}, in AIRR, image features $E_1(I_{cl})$ from our base encoder are fed into an attribute remover to learn an attribute-excluded representation $R(E_1(I_{cl}))$. The attribute remover is an $n$-layer convolutional block that is used to decorrelate the source attribute information from the image representation. A design requirement for the attribute remover is that it does not have skip connections since we aim to erase the attribute information from the encoded features, whereas skip connections would preserve this information.

To disentangle all the source attribute information from $E_1(I_{cl})$, we would need an attribute classifier to first identify these attributes. This can be achieved by Maximum Likelihood Estimation (MLE):
\begin{equation}
    {{L_d}\big(E_1(I_{cl})\big)} = -\sum\nolimits_{\mathbf{a_i} \in \mathbf{A}}{\mathbf{y_i}^T\log{p_c}(\mathbf{a_i}|E_1(I_{cl}))}
    \label{eq:mle}
\end{equation}
 where ${p_c}(\mathbf{a_i}|E_1(I_{cl}))$ is the probability distribution of $\mathbf{a_i}$, and $\mathbf{y_i}$ is the corresponding one-hot attribute label. We use one residual block as the attribute classifier to predict $p_c(\cdot)$.
 
After identifying the source attributes, we can then eliminate the attribute information in $R(E_1(I_{cl}))$ by minimizing their mutual information. Alternatively, it's easier to minimize the upper bound of this mutual information, which is the maximum log probability in the attribute class distribution added by a constant $c$ (See the Supplementary for proof of this upper bound):
 \begin{equation}
     \text{MI}\big({\mathbf{a}}_i,R(E_1(I_{cl}))\big) \leq c + {
     {\max}_{\mathbf{a}_i}} \log {p_c}\big(\mathbf{a}_i|R(E_1(I_{cl}))\big)
\end{equation}

Intuitively, minimizing this upper bound gives a uniform distribution over the attributes, meaning that the uncertainty for a specific attribute is maximized. Therefore, the
generated features would have little knowledge of what the
original attributes are. This loss function is thus defined as a margin loss:
 \begin{equation}
    {L_d}\big(R({E_1}({I_{cl}}))\big) = \sum\limits_{{{\mathbf{a}}_{\mathbf{i}}} \in {\mathbf{A}}} {\max \{} {
    {\max}_{\mathbf{a}_i}} \log {p_c}\big({{\mathbf{a}}_{\mathbf{i}}}|R(E_1(I_{cl}))\big) - \log  \frac{1}{{|{{\mathbf{a}}_{\mathbf{i}}}|}},c\},
    \label{eq:mim}
\end{equation}
where $|\mathbf{a_i}|$ is the number of attribute values in $\mathbf{a_i}$, and $c$ indicates the proximity to a uniform distribution, which is set to 0.01 in our experiments as we found it performed well. We refer to ${L_d}(R({E_1}({I_{cl}})))$ as a Mutual Information Minimization (MIM) loss.  Note that our attribute classifier and remover are trained end-to-end with our other generator and decoder components.

\subsection{Decoder}
\label{sec:decoder}
After the attribute remover, a new learned attribute representation is integrated with the disentangled features $R(E_1(I_{cl}))$ to generate the output image. Assuming a scale embedding vector $\bm{{\beta}_{a_i}}$ and a bias embedding vector $\bm{{\gamma}_{a_i}}$ for attribute $\mathbf{a_i}$, the original image $I$ thus can be reconstructed by combining $R\big(E_1(I_{cl})\big)$, $E_2(I_{bg})$ and the attribute embeddings, \ie,
\begin{equation}
    {I_{rec}} = G\Big(\text{concat}\big[\sum\nolimits_{\mathbf{a_i} \in \mathbf{A}} {{\bm{\beta _{\mathbf{a_i}}}} \cdot R\big({E_1}({I_{cl}})\big) + \bm{{\gamma _{\mathbf{a_i}}}}},E_2(I_{bg})\big] \Big),
\end{equation}
where $G$ is the decoder. Similarly, given target attributes of our desired output $\mathbf{\bar{A}}$, $R\big(E_1(I_{cl})\big)$ produces the manipulated image $I_{map}$ by
\begin{equation}
    {I_{map}} =G\Big(\text{concat}\big[\sum\nolimits_{\mathbf{\bar{a_i}} \in \mathbf{\bar{A}}} {{\bm{\beta} _{\mathbf{\bar{a_i}}}}} \cdot R\big({E_1}({I_{cl}})\big) + \bm{{\gamma} _{\mathbf{\bar{a_i}}}},E_2(I_{bg})\big]  \Big)
\end{equation}

In contrast to some prior work \cite{ak2019attribute,ChoiCKH0C18,kwak2020cafe}, where the attribute embedding vector and the encoded features are combined by concatenation, we multiply these two features such that linear interpolation between different attribute embeddings can better control the strength of these attributes in the output \cite{choi2020stargan,yang2021l2m}. Further discussion can be found in Section \ref{sec:abl}.

\subsection{Learning Objectives}
\label{sec:other_loss}
\noindent
\textbf{Disentanglement loss. } The disentanglement loss combines the MLE loss in Eq. (\ref{eq:mle}) and the MIM loss in Eq. (\ref{eq:mim}) as
\begin{equation}
    L_d={{L_d}\big(E_1(I_{cl})\big)}+{{L_d}\big(R\big(E_1(I_{cl})\big)\big)}
\end{equation}
By first identifying the source attributes in $E_1(I_{cl})$ and then minimizing the attribute information in $R(E_1(I_{cl}))$, this disentanglement loss enables the decoder to condition the output on the new attribute that is injected to the generator. In Section \ref{sec:abl}, we also show empirically that with the disentanglement loss, the attribute remover indeed gets rid of all the source attribute information.
\smallskip

\noindent
\textbf{Reconstruction loss. }The reconstructed image $I_{rec}$ is evaluated by its $l_1$ distance to the original image $I$:
\begin{equation}
    {L_{rec}} = ||{I_{rec}} - I||_1
\end{equation}

\noindent
\textbf{Adversarial loss.} The manipulated image $I_{map}$ doesn't have a paired ground truth of how it should look like, for which its plausibility is evaluated by the discriminator $D$. Using LSGAN \cite{mao2017least}, the adversarial loss of the generator can be written as
\begin{equation}
    {L_{adv}^{g}} = {(1 - D({I_{map}}))^2} + {(1 - D({I_{rec}}))^2}
\end{equation}

In the discriminator, this adversarial loss includes both the reconstructed image $I_{rec}$ and the manipulated image $I_{map}$ since they are both fake samples:
\begin{equation}
    L_{adv}^{d}=(1-D(I))^2+\frac{1}{2}\big(({D}(I_{map})^2+{D}(I_{rec})^2\big)
\end{equation}

\noindent
\textbf{Image attribute classification loss \cite{chen2016infogan}.} This loss maximizes the mutual information between the injected attribute embedding and the generated image:
\begin{equation}
\begin{aligned}
    L_{attr}^g
    &= - \sum\nolimits_{\mathbf{a_i} \in \mathbf{A}} {\mathbf{y_i}^T\log p_d(\mathbf{a_i}|I_{rec})}- \sum\nolimits_{\mathbf{\bar{a}_i} \in \mathbf{\bar{A}}} {\mathbf{\bar{y}_i}^T\log p_d(\mathbf{\bar{a_i}}|I_{map})}
\end{aligned}
\end{equation}
where $p_d(\cdot)$ is the probability distribution of attributes predicted by a classification branch in the discriminator. For the discriminator, this loss is defined on the real image $I$.
\begin{equation}
    L_{attr}^d=L_{attr}^d(I)= - \sum\nolimits_{\mathbf{a_i} \in \mathbf{A}} {\mathbf{y_i}^T\log p_d(\mathbf{a_i}|I)} 
\end{equation} 

\noindent
\textbf{Perceptual loss.} To further improve the quality of the generated images, a perceptual loss is introduced in the generator as in \cite{ak2019attribute}. It is based on the distance of paired real and fake images in the CNN feature space
\begin{equation}
    {L_p} = ||\text{CNN}(I) - \text{CNN}({I_{rec}})||_1 + ||\text{CNN}({I_{ref}}) -\text{CNN}({I_{map}})||_1
    \label{eq:lossp}
\end{equation}
where $I_{ref}$ is selected from the real images in the dataset to have exactly the same attributes $\mathbf{\bar{A}}$ as $I_{map}$.
\smallskip

\noindent
\textbf{Full objective.} Including all the above loss functions, the full objectives for the generator and discriminator are
\begin{equation}
    L_{gen}=L_{adv}^{g}+\lambda_1{L_d}+\lambda_2{L_{attr}^g}+\lambda_3{L_{rec}}+\lambda_4{L_p}
\end{equation}
\begin{equation}
    L_{dis}=L_{adv}^d+2\lambda_2 L_{attr}^d
\end{equation}
where $\lambda_1, \lambda_2, \lambda_3$ are trade-off parameters. As in prior work \cite{liu2021measuring,burns2021unsupervised}, these parameters must be set carefully to control the degree of disentanglement. Note that except for $L_{rec}$, these loss functions are all symmetrical with respect to $\mathbf{A}$ and $\mathbf{\bar{A}}$ to enforce that the manipulated image is a plausible reconstructed image.

\begin{table}[t]
    \centering
    \caption{Statistics of the datasets used in our experiments in Section \ref{sec:exp}}
     \scalebox{.8}[.8]{	
    \begin{tabular}{lccccc}\toprule
         \multirow{2}{*}{Dataset} &image &\#training &\#test &\multirow{2}{*}{\#attributes} &\#attribute \cr
         &size &images &images & &values  \cr \midrule 
         DeepFashion Synthesis \cite{liuLQWTcvpr16DeepFashion} &128x128 &76,979 &2000 &2 &21\cr
         DeepFashion Fine-grained Attribute \cite{liuLQWTcvpr16DeepFashion}  &{256x256} &{19,000} &{1000} &{6} &{26}\cr
         CelebA \cite{liu2015faceattributes} &128x128 &200,599 &2000 &8 &21\cr
         CelebA-HQ \cite{CelebAMask-HQ} &1024x1024 &29,000 &1000 &8 &21\cr
         \bottomrule
    \end{tabular}
    }
    \label{tab:dataset}
\end{table}
\section{Experiments}
\label{sec:exp}
To prove the efficiency of the proposed method, we evaluate our model on four publicly available datasets: DeepFashion Synthesis \cite{liuLQWTcvpr16DeepFashion}, DeepFashion Fine-grained Attribute \cite{liuLQWTcvpr16DeepFashion}, CelebA \cite{liu2015faceattributes} and CelebA-HQ \cite{CelebAMask-HQ}. On CelebA and CelebA-HQ, we group the attributes into 8 attribute categories and 21 attribute values following \cite{ak2021fashionsearchnet}. See Table \ref{tab:dataset} for detailed statics of each dataset.

\subsection{Implementation Details}

We use the model architecture in \cite{CycleGAN2017} as our backbone on DeepFashion and CelebA datasets. Following~\cite{yao2021latent}, on CelebA-HQ we adopt another backbone: StyleGAN2 \cite{karras2019style}, which is better suited to high resolution images. To improve training stability, we froze the weights of StyleGAN2's encoder and generator. Except for CelebA-HQ, the CNN used in Eq.\ (\ref{eq:lossp}) is a ResNet-50 model \cite{he2016deep} pretrained on image attribute classification task. On CelebA-HQ, we use StyleGAN2's encoder as the CNN in order to match the identity loss defined in \cite{yao2021latent}. 

For DeepFashion the target attributes are uniformly and randomly sampled from items in the same clothing category, \eg, dress and leggings. Here we did not sample the target attributes from the whole dataset because some annotated attributes can only appear in certain clothing categories. For example, leggings can't have V-neckline, and skirts can't have long sleeve. On CelebA-HQ, we traverse the values of the 8 attributes for each image during test for fair comparison with existing methods that use binary attribute values \cite{yang2021l2m,yao2021latent,shen2020interpreting}.

\subsection{Experimental Settings}

\noindent\textbf{Baselines.} We compare our model with related approaches on attribute manipulation: StarGAN \cite{ChoiCKH0C18}, AMNet \cite{ak2019attribute}, FLAM \cite{shin2019semi}, VPTNet \cite{kwon2022tailor}, AttGAN \cite{he2019attgan}, Student \cite{lezama2018overcoming}, FSNet-v2 \cite{ak2021fashionsearchnet}, CAFE-GAN \cite{kwak2020cafe}, InterfaceGAN \cite{shen2020interpreting}, L2M-GAN \cite{yang2021l2m} and LatentTransformer \cite{yao2021latent}. Among these approaches, FLAM, Student, InterfaceGAN and L2M-GAN also aim to achieve feature-level disentanglement, while AMNet, FSNet-v2 and CAFE-GAN aim to learn spatially disentangled representations. For StarGAN, AttGAN, Student, InterfaceGAN, L2M-GAN and LatentTransformer, we used the official implementations at author-provided links. AMNet and FLAM are reproduced by us following the configurations provided in the corresponding papers. Results of FashionSearchNet-v2, VPTNet and CAFE-GAN are directly copied from the original papers. For fair comparison on CelebA-HQ, we used StyleGAN2 encoded image features in InterfaceGAN.
\smallskip

\noindent\textbf{Evaluation metrics.} Following \cite{ak2019attribute,ak2021fashionsearchnet}, we use human evaluation and two standard metrics to evaluate the model's performance on attribute manipulation: attribute manipulation accuracy and top-k retrieval. Attribute manipulation accuracy, which is the classification accuracy of the target attribute on the manipulated images, measures the extent to which a model can modify the target attribute. We use a ResNet-50 model \cite{he2016deep} pretrained on attribute classification to evaluate the attribute manipulation accuracy on DeepFashion and CelebA. On CelebA-HQ, the accuracy is computed using the same facial attribute classifier as \cite{yao2021latent} for a fair comparison. Top-k retrieval, on the other hand, evaluates both the attribute changing and preservation capability. It is defined as the number of hits divided by the total number of queries. A query is called a hit if any of the manipulated image's top-k matches has exactly the target attributes in $\mathbf{\bar A}$. The top-k retrieval rate is averaged across all attributes. In all experiments, we use the deep features in the last fully-connected layer of the attribute classifier for image retrieval. All retrieval galleries have 20,000 images.

\subsection{Quantitative Results}
As shown in Table \ref{tab:deepfashion1}-\ref{tab:celebahq}, AIRR outperforms the state-of-the-art by a significant margin on most evaluation metrics. For example, Table \ref{tab:celebahq} shows the average attribute manipulation accuracy and top-k retrieval rates on CelebA-HQ, which boost performance by more than 20\% compared to existing methods. The improvements are more obvious on additive attributes, such as \emph{wearing hat}, which reports gains over prior work by more than 40\% in Table \ref{tab:celeba} and \ref{tab:celebahq}. Further discussion on ablations of our model can be found in Section \ref{sec:abl}.
\smallskip 

\begin{table}[!t]
    \centering
    \caption{Results on DeepFahison Synthesis. In our models, $\lambda_1=0.25, \lambda_2=0.125, \lambda_3=1.0, \lambda_4=1.0$}
    \scalebox{.65}[.65]{	
    \begin{tabular}{p{0.3\textwidth}>{\centering}p{0.1\textwidth}>{\centering}p{0.1\textwidth}>{\centering}p{0.1\textwidth}>{\centering}p{0.1\textwidth}>{\centering\arraybackslash}p{0.1\textwidth}} \toprule
    \multirow{2}{*}{Method} &\multicolumn{3}{c}{Manipulation Accuracy} &\multicolumn{2}{c}{Top-K Retrieval}  \cr \cmidrule(lr){2-4}   \cmidrule(lr){5-6}
    &Color &Sleeve &Avg. &R@5 &R@20 \cr \midrule
    StarGAN \cite{ChoiCKH0C18} &70.4 &77.2  &73.8  &71.1 &82.9\cr
    AMNet \cite{ak2019attribute}  &74.4  &82.1 &78.3  &85.1  &90.5 \cr
    AttGAN \cite{he2019attgan} &80.2 &91.0 &85.6 &90.6 &95.2\cr
    FLAM \cite{shin2019semi}  &- &-  &-  &26.7	&41.3\cr
    VPTNet \cite{kwon2022tailor} &- &85.7 &- &-&- \cr
    \midrule
    AIRR (w/o mask) &88.8 &92.2 &90.5 &94.4 &97.1\cr
    AIRR (w/o $L_d$) &93.9 &89.5 &91.7 &95.0 &97.3 \cr
    AIRR (w $L_h$) &89.4 &90.5 &90.0 &90.8 &92.4\cr
    AIRR    &\textbf{94.1}  &\textbf{96.5}  &\textbf{95.3}  &\textbf{97.6} &\textbf{98.8}  \cr
    \bottomrule
    \end{tabular}
    }
    \label{tab:deepfashion1}
\end{table}
\begin{table*}[!t]
    \centering
    \caption{Results on DeepFahison Fine-grained Attributes. In our models, $\lambda_1=0.05, \lambda_2=0.125, \lambda_3=2.0, \lambda_4=1.0$}
    \scalebox{.65}[.65]{	
    \begin{tabular}{p{0.3\textwidth}>{\centering}p{0.1\textwidth}>{\centering}p{0.1\textwidth}>{\centering}p{0.1\textwidth}>{\centering}p{0.1\textwidth}>{\centering}p{0.1\textwidth}>{\centering}p{0.1\textwidth}>{\centering}p{0.1\textwidth}>{\centering}p{0.1\textwidth}>{\centering\arraybackslash}p{0.1\textwidth}} \toprule
    \multirow{2}{*}{Method} &\multicolumn{7}{c}{Manipulation Accuracy} &\multicolumn{2}{c}{Top-K Retrieval}  \cr \cmidrule(lr){2-8}   \cmidrule(lr){9-10}
    &Pattern	&Sleeve	&Length	&Neckline	&Material	&Style &Avg. &R@5 &R@20 \cr \midrule
    StarGAN \cite{ChoiCKH0C18} &54.0	&38.7	&22.4	&44.3	&47.2	&24.6	&38.5   &25.1	&39.3\cr
    AttGAN \cite{he2019attgan} &47.4 &31.5 &19.0 &33.7 &40.3 &23.5 &32.6 &28.3 &39.1\cr
    AMNet \cite{ak2019attribute}  &53.6	&56.5	&24.4	&68.3	&47.9	&27.4	&46.4   &44.1	&47.5\cr 
     FLAM \cite{shin2019semi}   &- &-  &- &- &-  &- &- &17.6	&29.8  \cr\midrule
    AIRR (w/o mask) &70.7 &44.3 &19.0 &57.3 &55.0 &25.9 &45.4 &38.2 &52.4\cr
    AIRR (w/o $L_d$) &\textbf{89.1} &60.6 &31.8 &73.5 &74.9 &34.2 &60.8 &56.4 &68.7 \cr
    AIRR (w $L_h$) &85.1 &59.2 &25.7 &72.4 &72.6 &34.7 &58.2 &51.1 &60.2\cr
    AIRR    &87.8	&\textbf{65.9}	&\textbf{32.2}	&\textbf{74.5}	&\textbf{76.3}	&\textbf{35.8}	&\textbf{62.1} &\textbf{57.7}	&\textbf{70.3}\cr
    \bottomrule
    \end{tabular}
    }
    \label{tab:deepfashion2}
\end{table*}

\begin{figure*}[!t]
    \centering
    \begin{subfigure}[c]{0.25\textwidth}
    \centering
    \includegraphics[width=\textwidth]{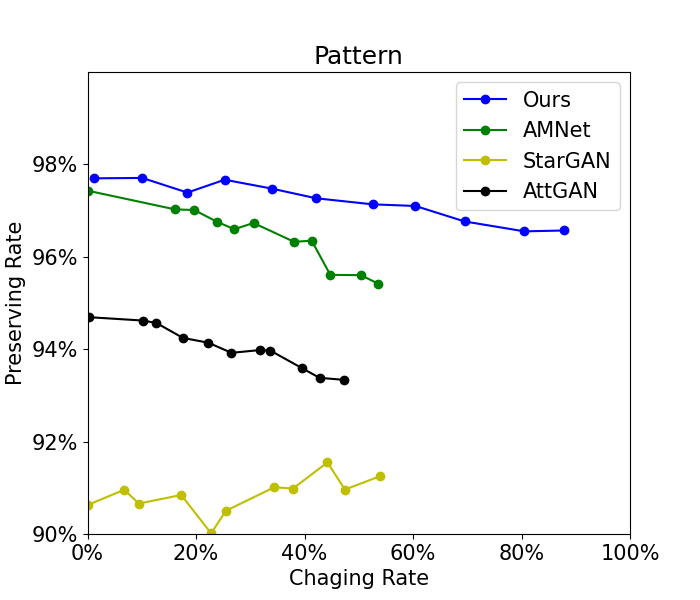}
    \end{subfigure}
    \begin{subfigure}[c]{0.25\textwidth}
    \centering
    \includegraphics[width=\textwidth]{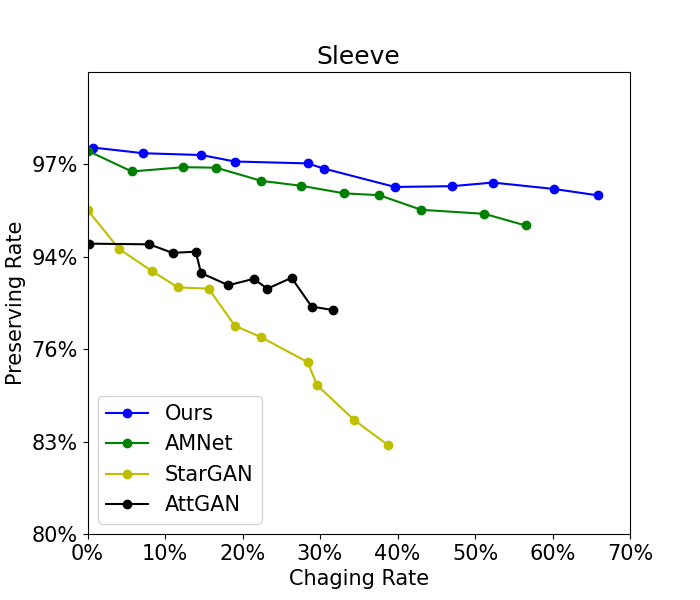}
    \end{subfigure}
    \begin{subfigure}[c]{0.25\textwidth}
    \centering
    \includegraphics[width=\textwidth]{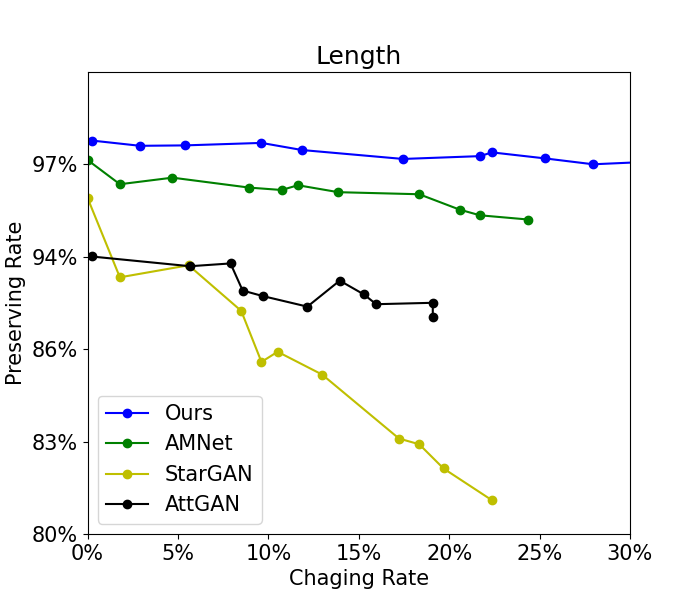}
    \end{subfigure}
    \begin{subfigure}[c]{0.25\textwidth}
    \centering
    \includegraphics[width=\textwidth]{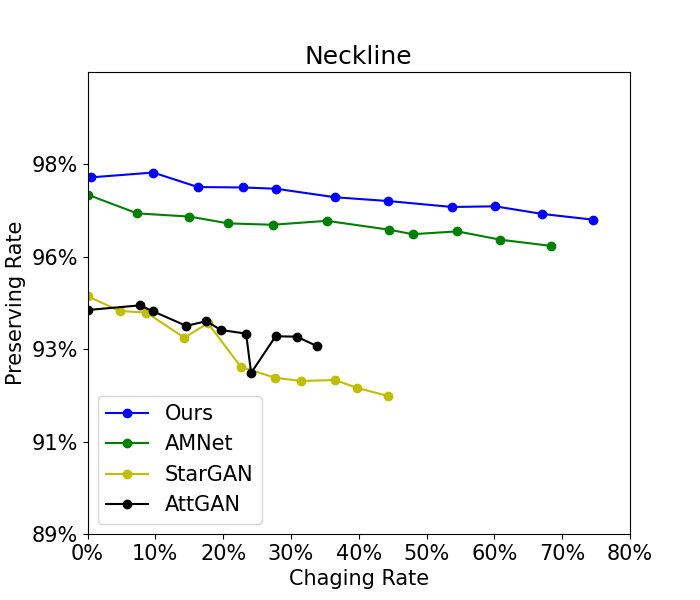}
    \end{subfigure}
    \begin{subfigure}[c]{0.25\textwidth}
    \centering
    \includegraphics[width=\textwidth]{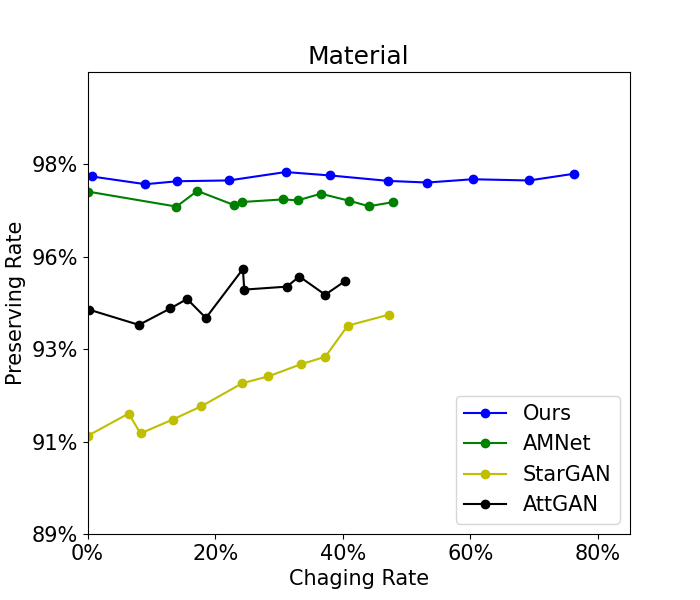}
    \end{subfigure}
    \begin{subfigure}[c]{0.25\textwidth}
    \centering
    \includegraphics[width=\textwidth]{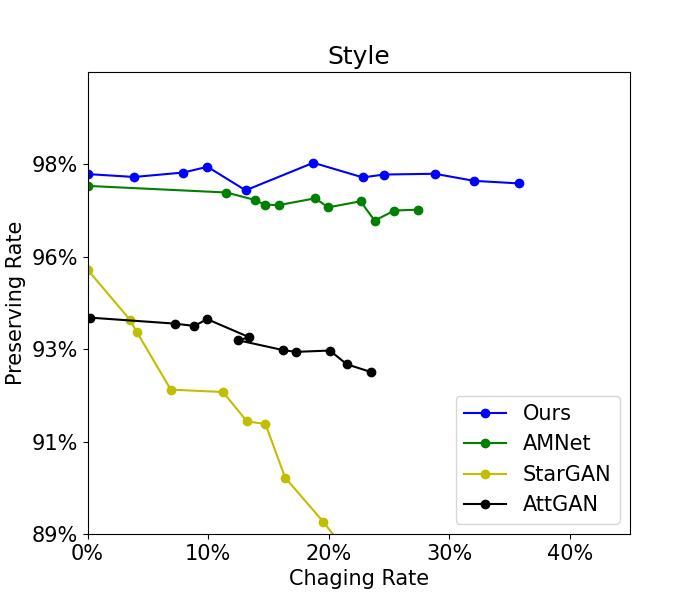}
    \end{subfigure}
    \caption{Attribute changing rate vs. attribute preservation rate. The interval of $y$ axis is made to be unequal for better visualization purposes}
    \label{fig:pc}
\end{figure*}

\noindent\textbf{Attribute preservation analysis. }To analyze what influence that changing a specific attribute has on preserving others, we gradually increase the ratio of manipulated images (\ie, the number of manipulated images divided by the number of all test images), and observe the ratio of successfully preserved attributes. Figure \ref{fig:pc} provides the attribute changing rate (\ie, attribute manipulation accuracy) vs. attribute preservation rate for each attribute in the DeepFashion dataset. In each graph, the preservation rates are averaged over all attributes excluding the target attribute. In Figure \ref{fig:pc}, our method achieves the highest preservation rate under the same attribute changing rate, proving its capability of controllable attribute manipulation as well as preservation.
\smallskip

\begin{table*}[!t]
    \centering
    \caption{Results on CelebA. In our models, $\lambda_1=0.5, \lambda_2=0.5, \lambda_3=1.0, \lambda_4=1.0$}
    \scalebox{.62}[.62]{	
    \begin{tabular}{p{0.25\textwidth}>{\centering}p{0.1\textwidth}>{\centering}p{0.1\textwidth}>{\centering}p{0.1\textwidth}>{\centering}p{0.1\textwidth}>{\centering}p{0.15\textwidth}>{\centering}p{0.1\textwidth}>{\centering}p{0.1\textwidth}>{\centering}p{0.1\textwidth}>{\centering}p{0.1\textwidth}>{\centering}p{0.1\textwidth}>{\centering\arraybackslash}p{0.1\textwidth}} \toprule
    \multirow{2}{*}{Method} &\multicolumn{9}{c}{Manipulation Accuracy} &\multicolumn{2}{c}{Top-K Retrieval}  \cr \cmidrule(lr){2-10}   \cmidrule(lr){11-12}
    &Hair Color	&Beard	&Hair Type	&Smiling	&Eyeglasses	&Gender	&Hat	&Age &Avg. &R@5 &R@20 \cr \midrule
    StarGAN \cite{ChoiCKH0C18} &55.2	&51.6	&35.6	&64.0	&86.1	&36.5	&6.4	&44.9	&38.8 &39.9	&55.2\cr
    Student \cite{lezama2018overcoming} &47.7 &43.5 &37.3 &60.0 &12.1 &42.7 &11.5 &39.8 &36.8 &38.2 &53.9\cr
    AMNet \cite{ak2019attribute}   &58.6	&34.4	&26.7	&43.7	&10.0	&21.0	&13.0	&22.4	&28.7 &33.1 &46.0\cr
    AttGAN \cite{he2019attgan} &72.6 &88.5 &48.1 &79.8 &94.7 &89.7 &21.7 &60.6 &69.5 &72.4 &86.5 \cr
    CAFE-GAN \cite{kwak2020cafe} &83.6 &40.1 &- &- &- &\textbf{95.2} &- &\textbf{88.6} &- &- &- \cr
    FSNet-v2 \cite{ak2021fashionsearchnet} &-   &-   &-   &-   &-   &-   &-   &-   &-   &68.0 &77.5\cr
     \midrule
    AIRR(w/o mask)    &\textbf{86.1}	&{96.0}	&\textbf{58.5}	&{92.7}	&{98.9}	&{91.3}	&{75.6}	&64.9	&{83.0} &{86.5}	&{94.3} \cr
    AIRR (w/o $L_d$) 
    &74.3 &93.9 &51.4 &92.4 &99.4 &89.5 &83.4 &69.7 &81.8 &87.1 &94.9\cr
     AIRR &75.9 &\textbf{96.4} &58.4 &\textbf{94.8} &\textbf{99.1} &91.6 &\textbf{93.1} &80.6 &\textbf{86.2} &\textbf{89.1} &\textbf{95.6}\cr
    \bottomrule
    \end{tabular}
    }
    \label{tab:celeba}
\end{table*}
\begin{table*}[!t]
    \centering
    \caption{Results on CelebA-HQ. In our models, $\lambda_1=0.25, \lambda_2=0.125, \lambda_3=20.0, \lambda_4=10.0$}
    \scalebox{.62}[.62]{	
    \begin{tabular}{p{0.3\textwidth}>{\centering}p{0.1\textwidth}>{\centering}p{0.1\textwidth}>{\centering}p{0.1\textwidth}>{\centering}p{0.1\textwidth}>{\centering}p{0.15\textwidth}>{\centering}p{0.1\textwidth}>{\centering}p{0.1\textwidth}>{\centering}p{0.1\textwidth}>{\centering}p{0.1\textwidth}>{\centering}p{0.1\textwidth}>{\centering\arraybackslash}p{0.1\textwidth}} \toprule
    \multirow{2}{*}{Method} &\multicolumn{9}{c}{Manipulation Accuracy} &\multicolumn{2}{c}{Top-K Retrieval}  \cr \cmidrule(lr){2-10}   \cmidrule(lr){11-12}
    &Hair Color	&Beard	&Hair Type	&Smiling	&Eyeglasses	&Gender	&Hat	&Age &Avg. &R@5 &R@20 \cr \midrule
   InterfaceGAN \cite{shen2020interpreting} &38.4 &\textbf{80.8} &36.4 &\textbf{97.7} &29.7 &55.8 &2.9 &42.0 &48.0 &19.1 &38.8\cr
   LatentTrans \cite{yao2021latent} &37.0	&78.8	&48.9	&85.3	&49.6	&62.2	&5.6	&47.4	&51.9	&20.7	&41.8\cr
    L2M-GAN \cite{yang2021l2m} &- &- &- &89.7 &- &- &- &- &- &- &-\cr
   \midrule
    AIRR(w/o mask+$L_d$)    &40.8 &73.1 &50.7 &93.2 &63.1 &71.4 &40.1 &83.3  &64.5 &51.3 &67.9\cr
    AIRR(w/o mask) &\textbf{54.8} &76.9 &\textbf{58.4} &95.4 &\textbf{88.2} &\textbf{79.0} &\textbf{49.2} &\textbf{88.0} &\textbf{73.7} &\textbf{60.9} &\textbf{75.5}\cr 
    \bottomrule
    \end{tabular}
    }
    \label{tab:celebahq}
\end{table*}

\begin{table}[!t]
\setlength{\tabcolsep}{1pt}
\centering
\begin{minipage}[t]{0.53\linewidth}
\centering
\caption{A/B user judgements for attribute manipulation correctness on the DeepFashion Fine-grained Attribute dataset}
\label{tab:human_fashion}
\scalebox{.9}[.9]{
\begin{tabular}{ccc}
\toprule
AIRR/StarGAN & AIRR/AttGAN & AIRR/AMNet \\
\midrule
65\%/35\% & 61\%/39\% & 54\%/46\%\\
\bottomrule
\end{tabular}
}
\end{minipage}\hfill%
\begin{minipage}[t]{0.43\linewidth}
\centering
\caption{A/B user judgements for attribute manipulation correctness on the CelebA-HQ dataset}
\label{tab:human_celeb}
\scalebox{.9}[.9]{
\begin{tabular}{cc}
\toprule
AIRR/InterfaceGAN & AIRR/LatentTrans \\
\midrule
76\%/24\% & 70\%/30\%\\
\bottomrule
\end{tabular}
}
\end{minipage}
\end{table}

\noindent\textbf{User study.} We also conducted human evaluation experiments on DeepFashion Fine-grained Attribute and CelebA-HQ using Amazon Mechanical Turk service to verify the quality of manipulated images. We tested on 50 images in each dataset, and different 5 worker were assigned per image. Each worker was presented 3 pictures: the original image, the manipulated image produced by AIRR, and the manipulated image generated by a randomly chosen baseline approaches in Table \ref{tab:human_fashion} or \ref{fig:celebahq}. The worker was asked to pick an image that better converts the specified attribute in the given image. Table \ref{tab:human_fashion} shows that 54-65\% of workers think our method achieves better attribute manipulation on the DeepFashion Fine-grained Attribute dataset. On CelebA-HQ, reported in Table~\ref{tab:human_celeb}, 70-76\% workers voted for AIRR, verifying its improved capability of manipulating facial attributes compared to prior work.
\begin{figure*}[!t]
    \centering
    \includegraphics[width=0.85\textwidth]{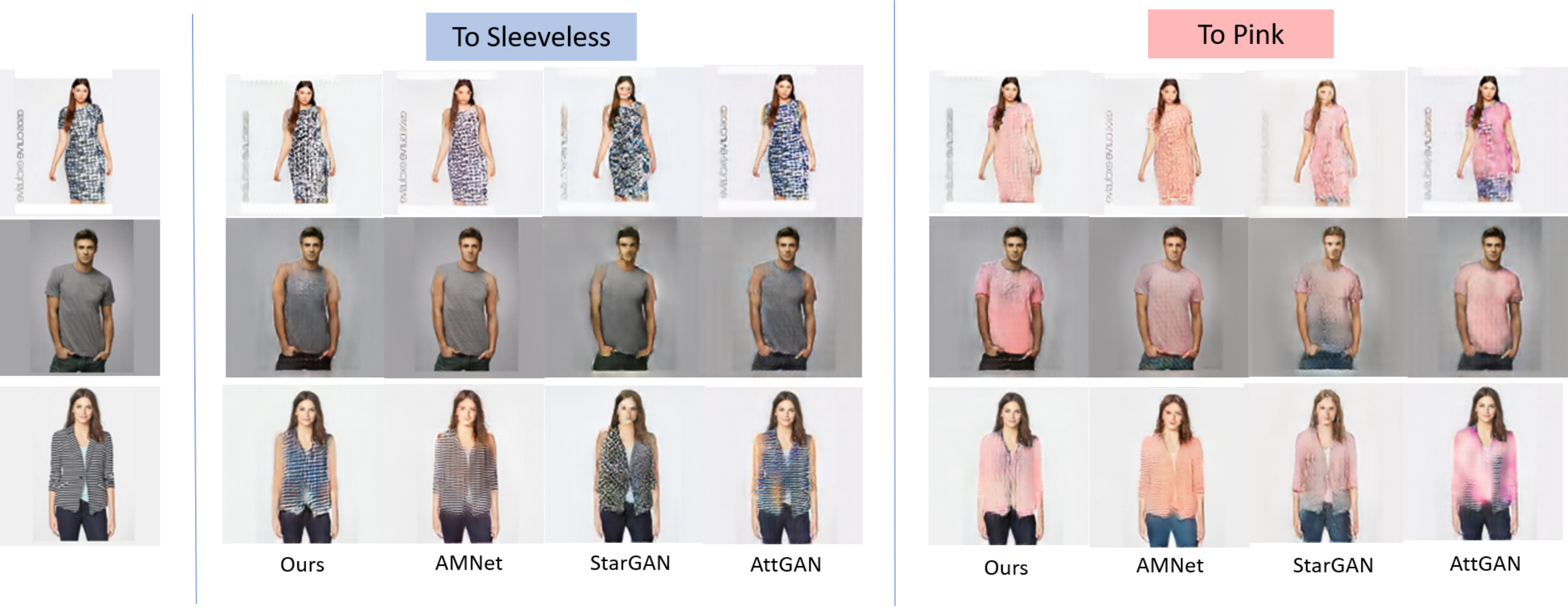}
    \caption{Qualitative examples on DeepFashion Synthesis. The first column shows original images}
    \label{fig:deepfashion1}
\end{figure*}
\begin{figure*}[!t]
    \centering
    \includegraphics[width=0.85\textwidth]{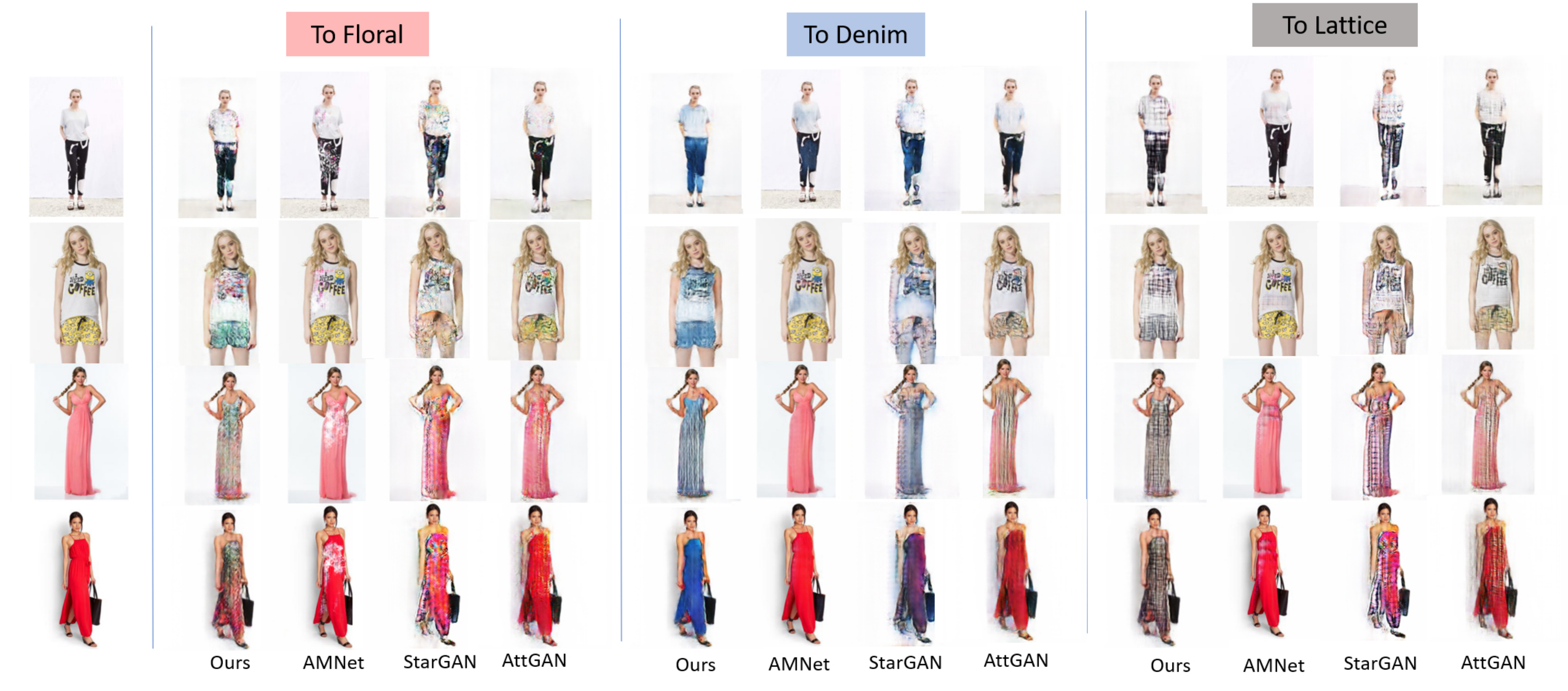}
    \caption{Qualitative examples on DeepFashion Fine-grained Attribute}
    \label{fig:deepfashion2}
\end{figure*}
\begin{figure*}[!t]
    \centering
    \includegraphics[width=0.9\textwidth]{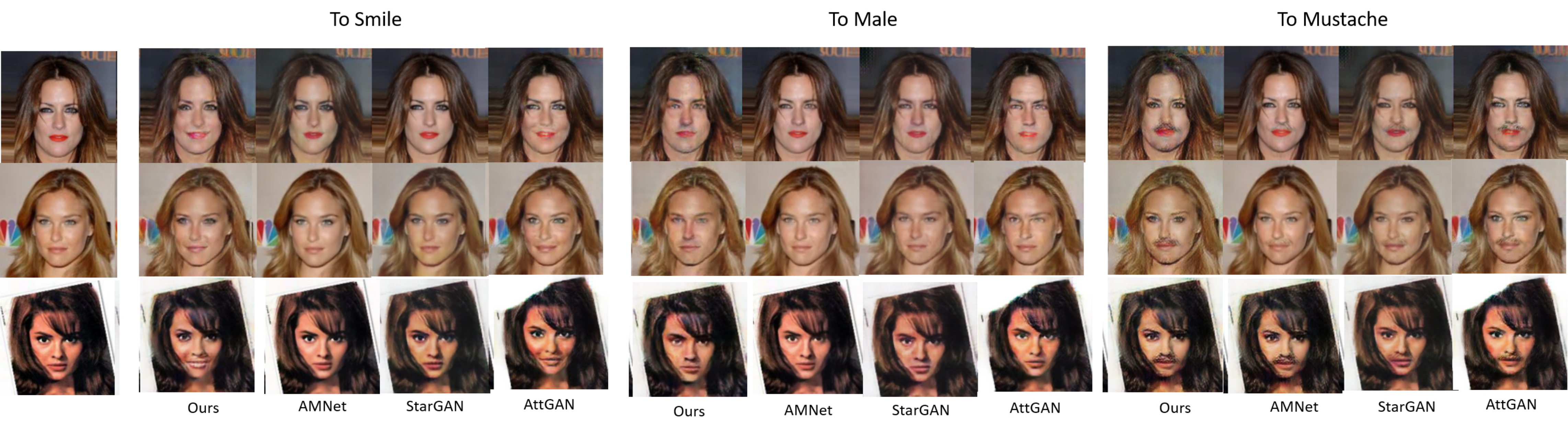}
    \caption{Qualitative examples on CelebA}
    \label{fig:celeba}
\end{figure*}

\subsection{Qualitative Results}
\begin{figure*}[!t]
    \centering
    \includegraphics[width=\textwidth]{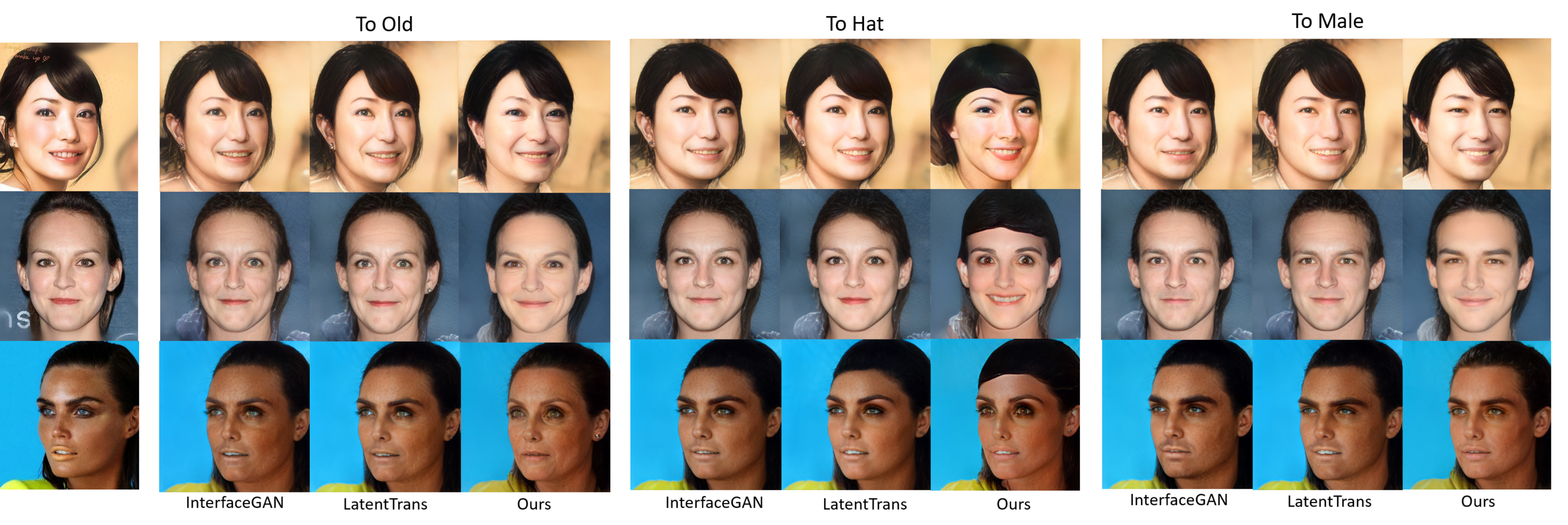}
    \caption{Qualitative examples on CelebA-HQ}
    \label{fig:celebahq}
\end{figure*}

Figure \ref{fig:deepfashion1}-\ref{fig:celebahq} presents some qualitative examples on each dataset that we used. For attributes that are relatively shallow and easy to learn, such as \emph{color} in Figure \ref{fig:deepfashion2}, all methods perform well in transforming the source attribute into the target attribute. Whereas for attributes with relatively more complicated semantics, \eg, \emph{lattice} in Figure \ref{fig:deepfashion2}, our method better represents the target attribute in the generated images. In addition, it can also be observed that due to the information hiding problem, several failures of previous methods exhibit as visually not altering the original image, such as AMGAN's \emph{To Denim} in Figure \ref{fig:deepfashion2} and LatentTrans's \emph{To Hat} in Figure \ref{fig:celebahq}. On the other hand, although our method avoids using the source attribute information for manipulation, it's failures can be more significant due to excessive manipulation. For example, the last two rows of \emph{To Floral} in Figure \ref{fig:deepfashion2} alter the dresses' appearance completely.

\subsection{Model Analysis}
\label{sec:abl}
\begin{figure*}[!t]
     \begin{subfigure}[c]{0.28\textwidth}
    \centering
    \includegraphics[width=\textwidth]{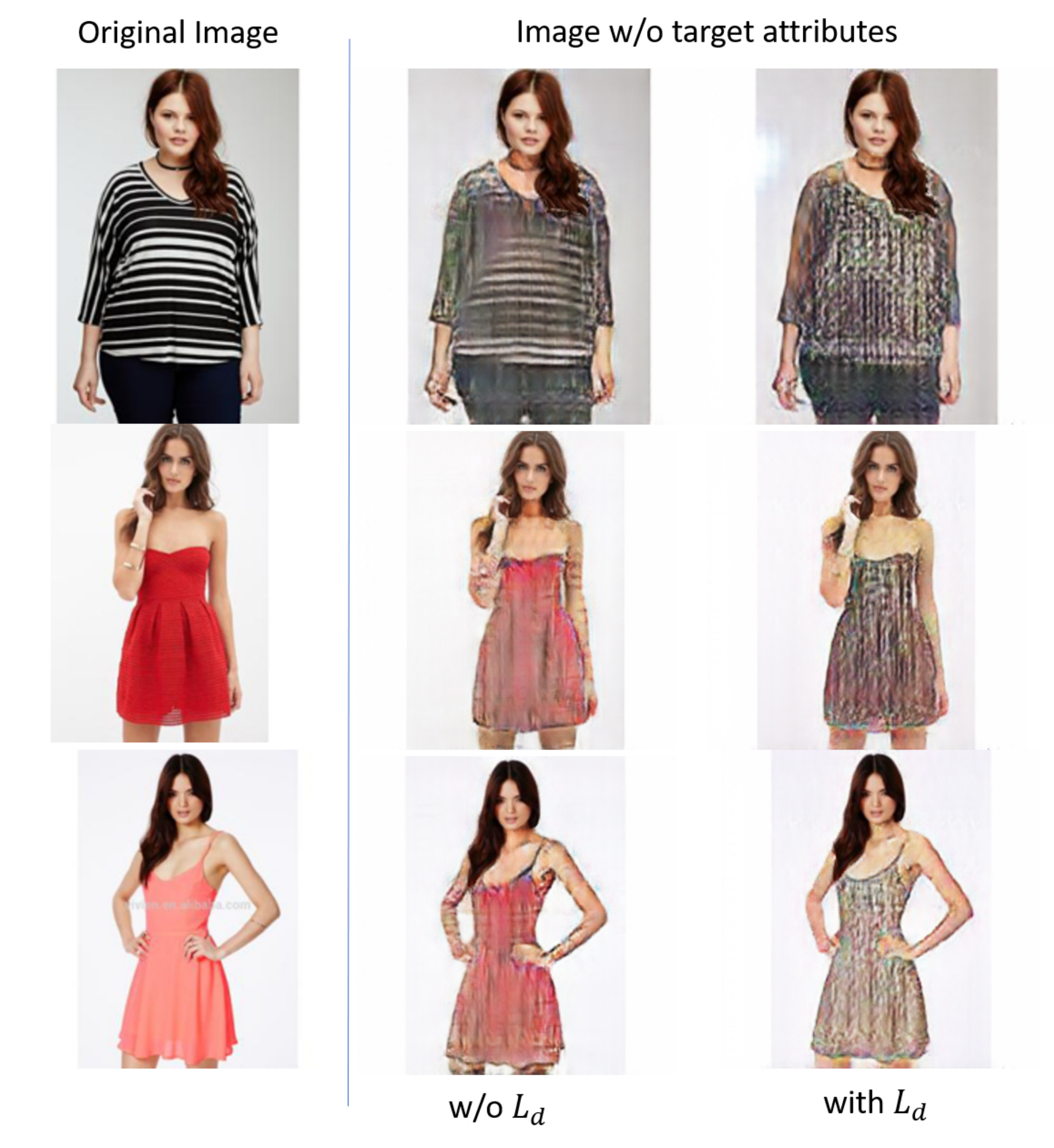}
     \caption{Examples on images without target attributes}
     \label{fig:wot}
    \end{subfigure}
    \hfill
    \begin{subfigure}[c]{0.35\textwidth}
    \centering
    \includegraphics[width=\textwidth]{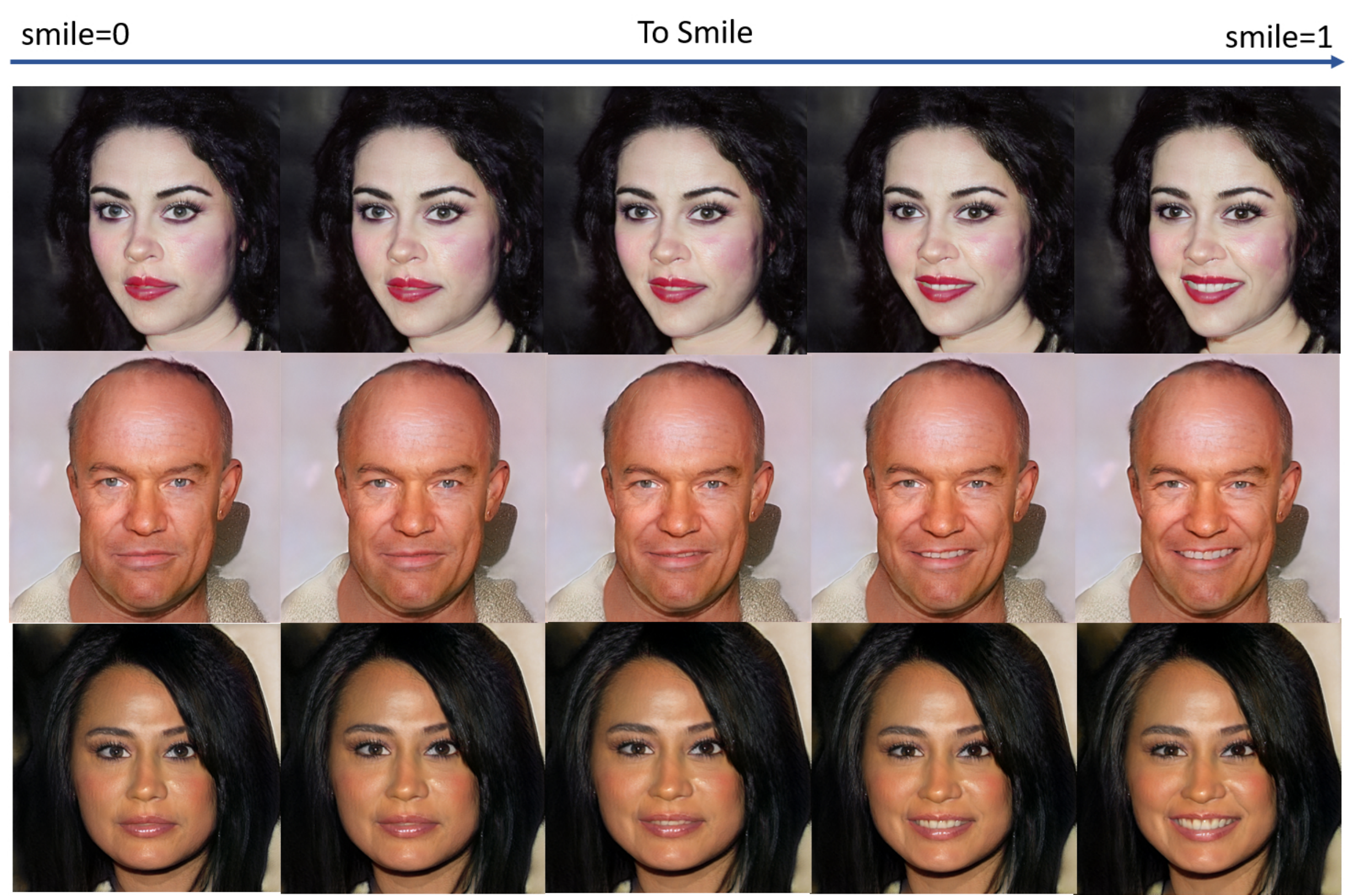}
     \caption{Examples on controlling attribute strength}
     \label{fig:beta}
    \end{subfigure}
    \hfill
     \begin{subfigure}[c]{0.3\textwidth}
    \centering
    \includegraphics[width=\textwidth]{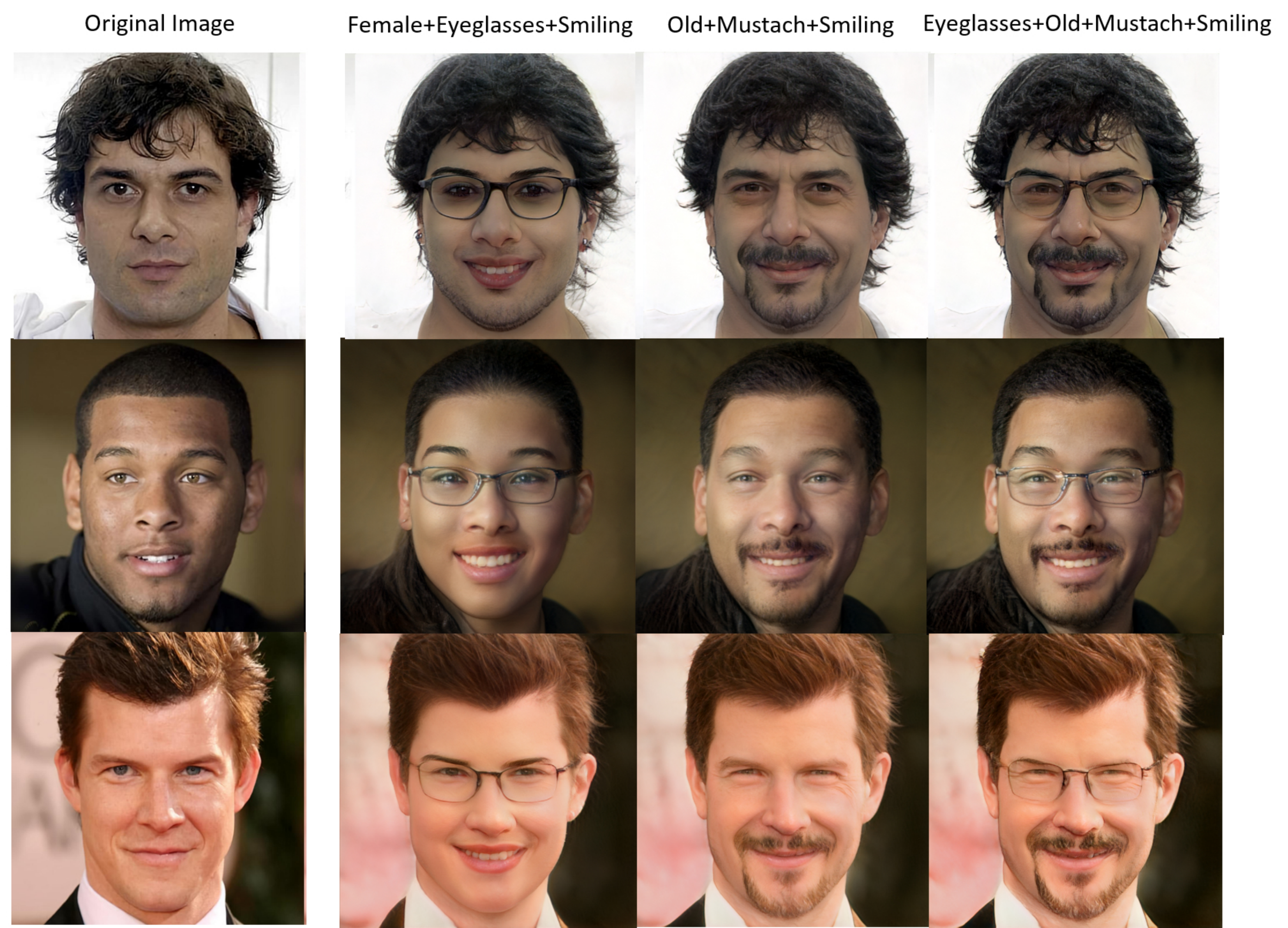}
     \caption{Examples on controlling multiple attributes}
     \label{fig:mattr}
    \end{subfigure}
    \caption{Visualizations used for model analysis in Sec.\ \ref{sec:abl}}
\end{figure*}

\noindent\textbf{Ablations of model components.} To demonstrate the contribution of the components of our proposed framework, we evaluate the performance of our model without the disentanglement loss and the parsing mask. Tables \ref{tab:deepfashion1}-\ref{tab:celebahq} report quantitative results of ablations of our model. AIRR (w/o $L_d$), which disables the attribute remover, causes losses on average attribute manipulation accuracy and top-5 retrieval on all four datasets. Especially in CelebA-HQ, AIRR (w/o $L_d$) losses 9\% accuracy compared to AIRR. This suggests that disentangling attribute information by decorrelation is effective in image manipulation. We also explored what the generated images would look like when injecting no attribute, \ie, setting the target attribute's scale embedding vector to be $\mathbf{1}$ and the bias embedding vector to be $\mathbf{0}$. This way we can check if the model is hiding source attribute information in the encoded features. If it suffers from information hiding, then the source attributes should been seen in generated images. In Figure \ref{fig:wot}, without the proposed disentanglement loss, most source attributes, including material and pattern, indeed appear in the generated images. With the proposed attribute excluded representation, all these attribute information is successfully removed, avoiding the information hiding problem suffered by prior work. 

In the meanwhile, AIRR (w/o mask), which removes the parsing mask along with the second encoder, also degrades the evaluation metrics as seen in TAbles~\ref{tab:deepfashion1}-\ref{tab:celebahq}. This indicates the importance of concentrating manipulation on the object of interest.  However, we note that even without the parsing mask, our approach still outperforms prior work on most metrics. Note that in the two ablations of CelebA-HQ, we didn't add the parsing mask in order to reduce the computational costs for generating high resolution images.

We also tried replacing the proposed disentanglement loss with the honesty loss in \cite{bashkirova2019adversarial}, which was introduced to avoid the general information hidden problem when using cycle consistency. In Table \ref{tab:deepfashion1} and \ref{tab:deepfashion2}, AIRR still outperforms AIRR (w $L_h$) that adopts the honesty loss, suggesting that the proposed disentanglement loss is more targeted to the attribute manipulation task.  See the Supplementary for more ablation results for hyperparameters used by our model.
\smallskip

\noindent\textbf{Interpolation of attribute values. } Linear interpolating between different attribute embedding vectors $\bm{\beta_{a_i}}$ and $\bm{\gamma_{a_i}}$ corresponds to an interpolation between different values of the target attribute. Take "smile" for example, Figure \ref{fig:beta} gives the outputs of interpolating from the \emph{not smiling} embedding vector to the \emph{smiling} embedding vector. Let $c$ be the weight (\ie, interpolation coefficient) of the target attribute \emph{smiling}. The smile in generated images gradually builds up as $c$ increases, showing a continuous control over the attribute strength.
\smallskip


\noindent\textbf{Controlling multiple attributes in one forward pass. } In some prior work, \eg, \cite{yao2021latent,wang2021attribute,yang2021l2m}, multi-attribute editing is often accomplished by sequential manipulation, \ie, edit one attribute at a time. In contrast, AIRR is capable of changing multiple attributes in a single forward-pass by directly specifying the input target attributes in $\mathbf{\bar A}$. Figure \ref{fig:mattr} gives some examples on CelebA-HQ. Even manipulating 3 or 4 attributes at the same time, our model is able to edit only the specified attributes without influencing other information in the image.

\section{Conclusion}
In this paper, we propose Attribute Information Removal and Reconstruction (AIRR) network for image editing. The attribute information removal and reconstruction module in AIRR produces an attribute excluded representation, eliminating sources of information hiding suffered by prior work. Results on four diverse datasets including DeepFashion Synthesis, DeepFashion Fine-grained Attribute, CelebA and CelebA-HQ, report that our model improves attribute manipulation accuracy and top-k retrieval rate by 10\% on average over prior work. A user study also demonstrates that images with attributes manipulated with our approach are preferred in up to 76\% of cases. One direction for future work is to explore controllable attribute manipulation in unsupervised setting.

\noindent\textbf{Acknowledgements.} This material is based upon work supported, in part, by DARPA under agreement number HR00112020054 and the National Science Foundation under Grant No.\ DBI-2134696. Any opinions, findings, and conclusions or recommendations expressed in this material are those of the author(s) and do not necessarily reflect the views of the supporting agencies.

%
%
\bibliographystyle{splncs04}
\bibliography{mybib}

\newpage
\centerline{\huge{\textbf{Supplementary}}}

\setcounter{section}{0}
\section{Proof of the Upper Bound for Mutual Information}
In Eq. (2) of the paper, we claimed that the mutual information between the source attributes ${{\mathbf{a}}_i}$ and the attribute excluded features $R(E_1(I_{cl}))$ is upper bounded by the maximum log probability in the attribute distribution. We prove this claim in the following.

Let $\text{MI}({{\mathbf{a}}_i},R(E_1(I_{cl})))$ denote the mutual information. Replacing $R(E_1(I_{cl}))$ with $r$ for convenience gives
\begin{equation} \tag{15}
\begin{aligned}
  {\text{MI}}({{\mathbf{a}}_i},r) &= \sum\nolimits_r {\sum\nolimits_{{{\mathbf{a}}_i}} {p({{\mathbf{a}}_i},r)\log \frac{{p({{\mathbf{a}}_i},r)}}{{p({{\mathbf{a}}_i})p(r)}}} }  \\
   &= \sum\nolimits_r {\sum\nolimits_{{{\mathbf{a}}_i}} {p({{\mathbf{a}}_i},r)\log \frac{{p({{\mathbf{a}}_i}|r)}}{{p({{\mathbf{a}}_i})}}} } \\
   &=\sum\nolimits_r {\sum\nolimits_{{{\mathbf{a}}_i}} {p({{\mathbf{a}}_i},r)[\log p({{\mathbf{a}}_i}|r) - \log p(} } {{\mathbf{a}}_i})]
   \end{aligned}
   \end{equation}  
Since the number of attribute values in ${{\mathbf{a}}_i}$ is finite, $-\log p(\mathbf{a}_i)$ can be upper bounded by a constant $c,c>0$:
\begin{equation} \tag{16}
\begin{aligned}
     {\text{MI}}({{\mathbf{a}}_i},r) 
   &\leq \sum\nolimits_r {\sum\nolimits_{{{\mathbf{a}}_i}} {p({{\mathbf{a}}_i},r)\log p({{\mathbf{a}}_i}|r) + c\sum\nolimits_r {\sum\nolimits_{{{\mathbf{a}}_i}} {p({{\mathbf{a}}_i},r)} } } }  \hfill \\
   &= \sum\nolimits_r {\sum\nolimits_{{{\mathbf{a}}_i}} {p({{\mathbf{a}}_i},r)\log p({{\mathbf{a}}_i}|r) + c} }  \hfill \\ 
\end{aligned} 
\end{equation}  
 In the r.h.s., we can continue upper bounding ${{p({{\mathbf{a}}_i}|r)}}$ with the maximum probability in the distribution to make it independent of ${\mathbf{a}}_i$:
    \begin{equation} \tag{17}
    \begin{aligned}  
   {\text{MI}}({{\mathbf{a}}_i},r) &\leq \sum\nolimits_r {\max \limits_{{{\mathbf{a}}_i}}  \log p({{\mathbf{a}}_i}|r)\sum\nolimits_{{a_i}} {p({{\mathbf{a}}_i},r)} }+c  \\
   &= \sum\nolimits_r {p(r)\max \limits_{{{\mathbf{a}}_i}}  \log p({{\mathbf{a}}_i}|r)}+c  \\ 
   &= {\mathbb{E}_{r \sim {p(r)}}}[ \max \limits_{{{\mathbf{a}}_i}}  \log p({a_i}|r) ]+c, \\ 
    \end{aligned}
\end{equation}
where $c$ is a constant. Note that we can not minimize the mutual information itself because the joint distribution $p(\mathbf{a}_i,r)$ is intractable. The tightness of this upper bound depends on the distribution $p(\mathbf{a}_i)$ and $p(\mathbf{a}_i|r)$. More specifically, larger $\min _{\mathbf{a_i}}p(\mathbf{a_i})$ gives smaller constant $c$, and smaller $\max _{\mathbf{a_i}}p(\mathbf{a_i}|r)$ reduces the gap. The equality is reached when $p(\mathbf{a_i}|r)$ is an uniform distribution.

To conclude, using an attribute classifier to estimate the above conditional probability $p({a_i}|r)$, we prove that the upper bound is the maximum log probability in the attribute distribution as in Eq. (2).
\section{Ablations on Hyperparameters}
\begin{figure} 
    \centering
    \vspace{-6mm}
    \begin{subfigure}[c]{0.9\textwidth}
    \centering
    \includegraphics[width=0.45\textwidth]{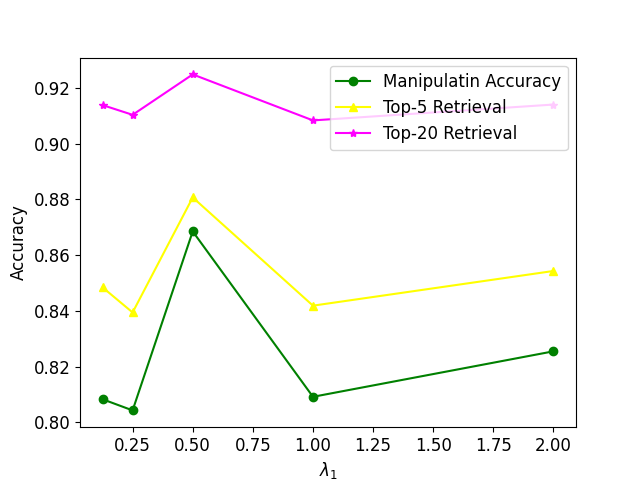}\quad
    \includegraphics[width=0.45\textwidth]{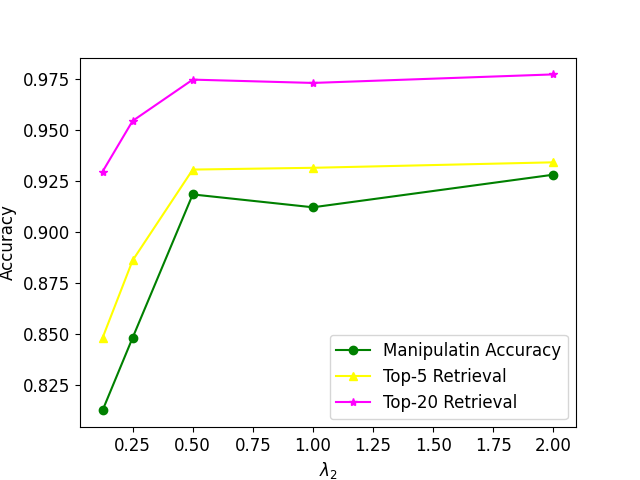}\\
    \includegraphics[width=0.45\textwidth]{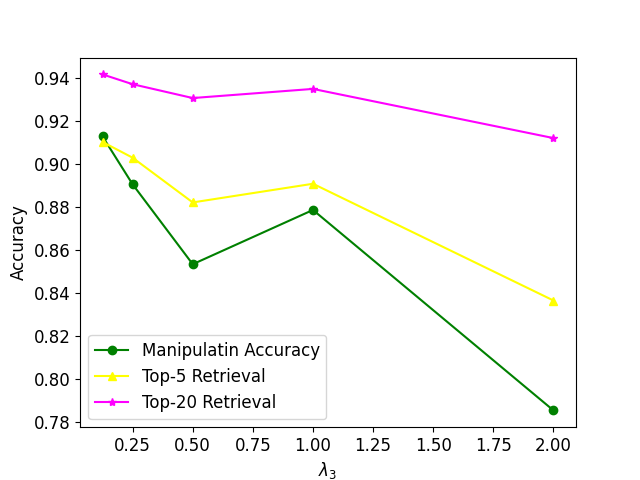}\quad
    \includegraphics[width=0.45\textwidth]{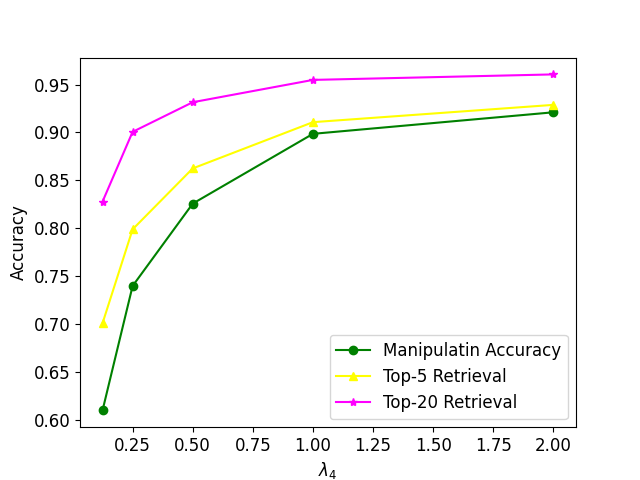}
    \caption{Parameter value v.s. Accuracy. Higher is better}
    \label{fig:lambdasa}
    \end{subfigure}
    \vfill
    \begin{subfigure}[c]{0.4\textwidth}
    \centering
    \includegraphics[width=\textwidth]{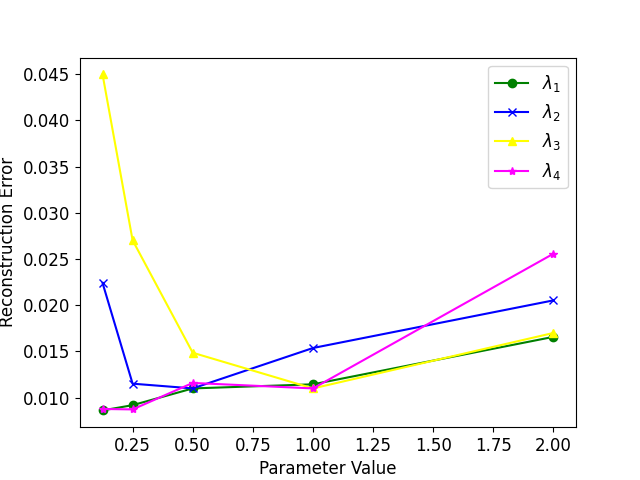}
    \caption{Parameter Value v.s. Reconstruction Error. Lower is better}
    \label{fig:lambdasb}
    \end{subfigure}
    \caption{Results on using different values of the hyperparameters. $\lambda_1$ to $\lambda_4$ denotes the trade-off parameters for disentanglement, image attribute prediction, image reconstruction and perceptual loss, respectively}
    \label{fig:lambdas}
\end{figure}

In Figure \ref{fig:lambdas}, we provide the experimental results for setting different values of the hyperparameters in Eq. (13) and (14) on CelebA. $\lambda_1$ to $\lambda_4$ denotes the trade-off parameter for disentanglement, image attribute prediction, image reconstruction and perceptual loss, respectively. Figure \ref{fig:lambdasa} shows the manipulation accuracy, top-5 retrieval and top-20 retrieval rates for each parameter. The reconstruction error has a different unit of measurement, for which we show its corresponding graph in Figure \ref{fig:lambdasb}. It can be noticed that increasing the weight (\emph{i.e.}, $\lambda_2$) for the image attribute loss improves the manipulation accuracy, whereas it can hurt the reconstruction performance. This indicates a trade-off between successful manipulation and qualitative reconstruction. In the paper, we chose the values of each trade-off parameter for a balance between these two aspects.

\begin{figure} 
    \centering
    \begin{subfigure}[c]{\textwidth}
    \centering
    \includegraphics[width=\textwidth]{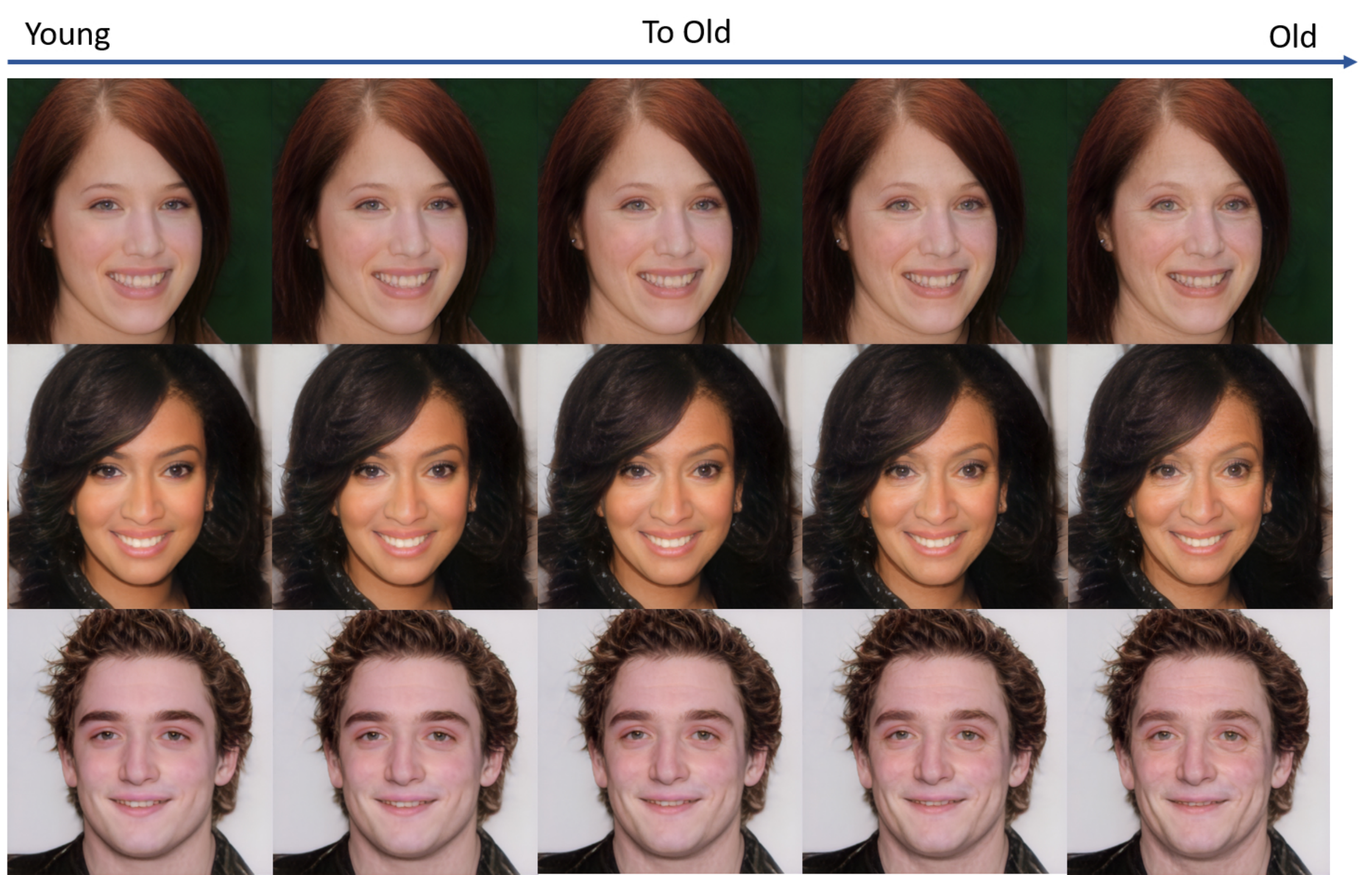}
    \end{subfigure}
    \vfill
    \begin{subfigure}[c]{\textwidth}
    \centering
    \includegraphics[width=\textwidth]{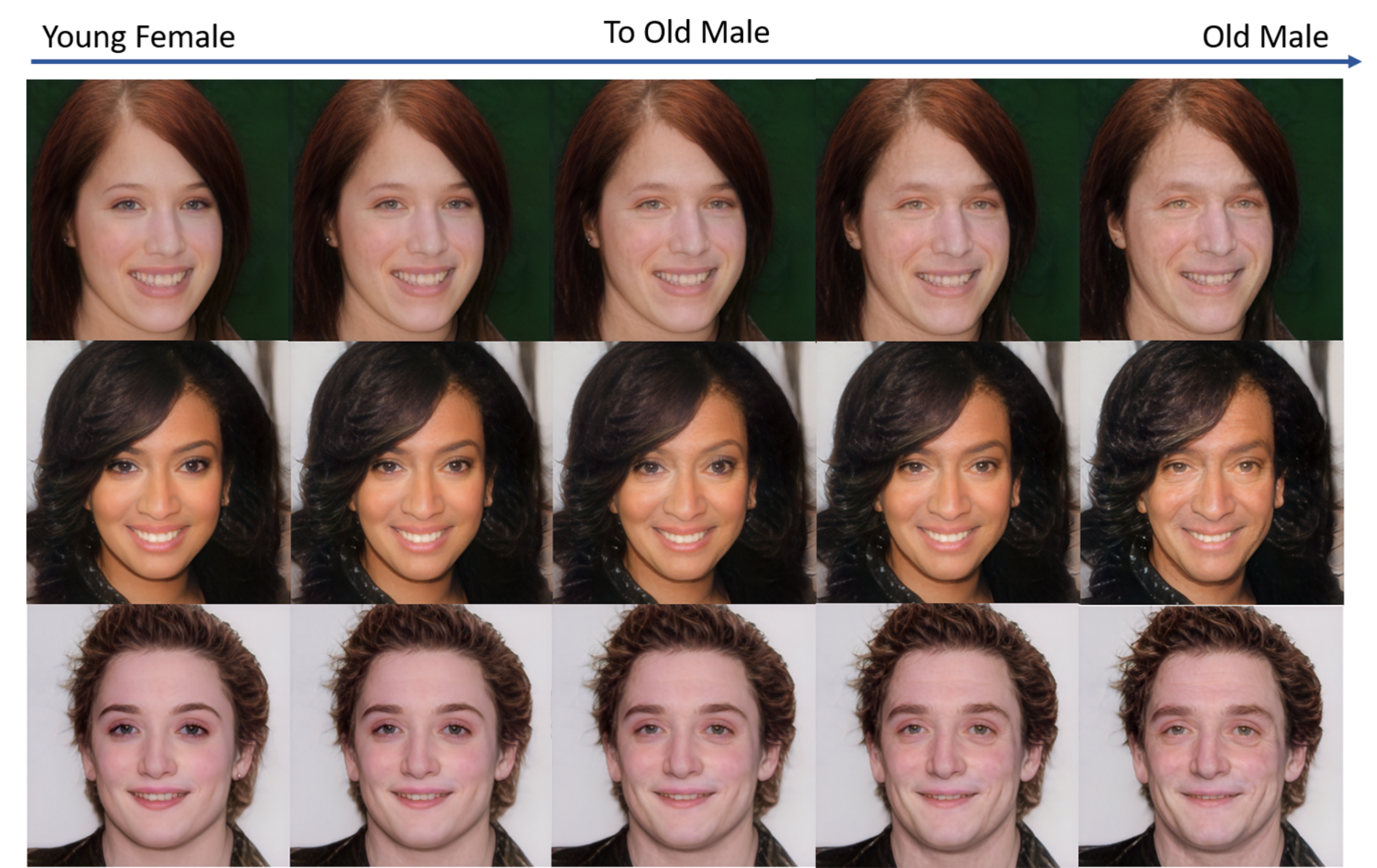}
    \end{subfigure}
    \caption{Additional examples on manipulating the attribute strength}
    \label{fig:more_attr_str}
\end{figure}
\end{document}